\pdfoutput=1
\documentclass[10pt,twocolumn,letterpaper]{article}

\usepackage{cvpr}              % To produce the CAMERA-READY version
\usepackage{graphicx}
\usepackage{amsmath}
\usepackage{amssymb}
\usepackage{booktabs}
\usepackage[dvipsnames,table,xcdraw]{xcolor}

\usepackage[pagebackref,breaklinks,colorlinks]{hyperref}

\usepackage[capitalize]{cleveref}
\crefname{section}{Sec.}{Secs.}
\Crefname{section}{Section}{Sections}
\Crefname{table}{Table}{Tables}
\crefname{table}{Tab.}{Tabs.}

\def\cvprPaperID{*****} % *** Enter the CVPR Paper ID here
\def\confName{CVPR}
\def\confYear{2022}

\begin{document}

\title{MSTRIQ: No Reference Image Quality Assessment Based on Swin Transformer with Multi-Stage Fusion}

\author{Jing Wang\footnotemark[1], Haotian Fan\footnotemark[1], Xiaoxia Hou\footnotemark[1], Yitian Xu, Tao Li, Xuechao Lu, Lean Fu\\
ByteDance Inc\\
Building 10, Zone A, Business Park, Lane 888, Tianlin Road, Minhang District, Shanghai\\
{\tt\small \{wangjing.crystalw, fanhaotian, houxiaoxia, xuyitian.mars, litao.walker, luxuechao, fulean\}}\\
{\tt\small @bytedance.com}
%\and
%Haotian Fan\\
%{\tt\small fanhaotian@bytedance.com}
%\and
%Xiaoxia Hou\\
%{\tt\small houxiaoxia@bytedance.com}
}
\maketitle
\renewcommand{\thefootnote}{\fnsymbol{footnote}}
\footnotetext[1]{These authors contributed equally to this work}
%%%%%%%%% ABSTRACT
\begin{abstract}
Measuring the perceptual quality of images automatically is an essential task in the area of computer vision, as degradations on image quality can exist in many processes from image acquisition, transmission to enhancing. Many Image Quality Assessment(IQA) algorithms have been designed to tackle this problem. However, it still remains unsettled due to the various types of image distortions and the lack of large-scale human-rated datasets. In this paper, we propose a novel algorithm based on the Swin Transformer\cite{31} with fused features from multiple stages, which aggregates information from both local and global features to better predict the quality. To address the issues of small-scale datasets, relative rankings of images have been taken into account together with regression loss to simultaneously optimize the model. Furthermore, effective data augmentation strategies are also used to improve the performance. In comparisons with previous works, experiments are carried out on two standard IQA datasets and a challenge dataset. The results demonstrate the effectiveness of our work. The proposed method outperforms other methods on standard datasets and ranks 2nd in the no-reference track of NTIRE 2022  Perceptual Image Quality Assessment Challenge\cite{53}. It verifies that our method is promising in solving diverse IQA problems and thus can be used to real-word applications.
\end{abstract}

%%%%%%%%% BODY TEXT
\section{Introduction}
\label{sec:intro}

Image quality assessment aims to quantitatively evaluate the perceptual quality of images\cite{1,2,3}. It has wide applications in many areas such as image accuisition, compression and processing\cite{4,5,6}. Owing to its important roles in these areas, image quality assessment has gained increasing research intereset during the years. 

General methods of assessing image quality can be mainly classified into two categories: subjective rating and objective metrics. Subjective rating measures image quality by Mean Opinion Score(MOS), where a group of people will be asked to rate a series of images according to their visual quality. The quality of each image is then represented by the average of human ratings. Although subjective scores are accurate, collecting them is time-consuming and complicated. It requires a lot of people to participate in rating each image, and these people will need to be well-trained. Research show that the subjective scores can be influenced by a great range of aspects, including the number of participants, watching distances and display devices\cite{7}, thus they are not practical to be used real-time and large-scale IQA tasks. 

As alternatives, objective metrics employ computer vision techniques to automatically learn the relationship between tested images and human perceived quality\cite{8,9,10,11,12}. The predicted results of these methods are expected to be equal to human ratings. According to the availability of original pristine-quality images, there are different kinds of objective IQA algorithms that are used in corresponding situations: no-reference(NR), full-reference(FR) and reduced-reference(RR) algorithms\cite{13,14,15}. In this paper, we are discussing the NR-IQA algorithms, which do not require references from original pristine-quality images and thus can be used when distorted images are the only inputs. 

In the past few years, traditional NR-IQA algorithms such as BRISQUE\cite{16}and NIQE\cite{17}are put forward to solve this problem. These traditional algorithms have implemented handcrafted features of image stastitics or learn with codebooks to predict image quality. Research found that performances of traditional algorithms can compare with human ratings in assessing the quality of images generated in well-controled laboratory environmnets, but are not satisfying when employed in real-world applications\cite{10}. With the advancement of deep learning techniques in computer vision, algorithms based on neural networks are widely developed\cite{18}. Convolutional Neural Network(CNN) such as AlexNet\cite{19} and ResNet\cite{20} are proposed. Well-designed CNNs can extract both global and local visual features, which prove to be effective in many fields of computer vision, including  image classification, detection, segmentation and so on\cite{19,20,21}. Research in \cite{22} is the first work intergrating CNN into IQA tasks. This research is then greatly extended by rankIQA\cite{23}, DBCNN\cite{24}, hyperIQA\cite{25} and so on. 

After this, the vision transformers(ViTs) emerge as competitive alternatives to CNNs, which outperform the preceding CNN architecture in the mentioned vision tasks\cite{26,27,28}. TRIQ\cite{29}, MUSIQ\cite{30} are previous works that intergrate ViTs into no-reference image quality assessment. These works show that the vision transformers have a lot of potential for addressing the problem. Different from the ViTs in pioneering works, our proposed method is based on the transformer with hierarchy architecture and shifted windows(Swin Transformer)\cite{31, 32}. We use it as the backbone because it inherits the advantages of both CNNs and previous ViTs, which achieves a trade-off between inference speed and accuracy. To evaluate the distortions of different scales, a multi-stage fusion strategy is adopted to aggregate information. Features from intermediate layers are of vitality to IQA tasks. We also employ data augmentation and training strategies to improve performance. Experiments are carried out to compare the proposed method with pioneering works. The results prove the superiority of our method in terms of both SRCC and PLCC on two standard IQA datasets. The proposed algorithm also ranked 2nd in the no-reference track of NTIRE 2022  Perceptual Image Quality Assessment Challenge. We will discuss the details about the challenge later.

In the subsequent sections, this paper will be organized as follows. In Section 2, related works will be introduced. In Section 3 and 4, details of the proposed method and experimental results will be demonstrated. Conclusions will be summarized in the last section.

%-------------------------------------------------------------------------

\section{Related work}
\subsection{Image quality assessment}
Research about image quality assessment attempts to develop a method that is capable of predicting perceptual quality of images automatically without human participating\cite{3}. There is an increasing disire of IQA techniques in the image processing systems nowadays, as the visual content of images can undergo a various types of distortions\cite{33,34} in every part of the systems. Image quality assessment techniques help optimize the steps in image processing, by evaluating their outputs and giving a feedback on how the quality is changing. This can be done by subjective ratings. However, accurate subjective ratings are restricted to only particular circumstances, which arouses extensive research about objective metrics\cite{2}. Early objective IQA algorithms mainly use handcrafted features or learn with codebooks\cite{16, 17, 35,36,37}. In the wake of developments in deep learning techniques\cite{18,19,20}, the employment of neural networks further improves the performance of IQA algorithms to a new stage\cite{25, 38, 39}. CNNs, vision transformers and combinations of both become frequently used backbones. The differences of these approaches will be discussed in the later sections.

\subsection{Learning to rank}

One of the most common loss functions for training objective NR-IQA models is regression loss such as Mean Squared Error(MSE) and Mean Absolute Error(MAE), which measures the difference between the predicted quality score and the ground truth for each image\cite{40}. However, regression loss alone is not enough for training deep learning models if there is no sufficient labeled data\cite{7}. As mentioned before, obtaining large-scale IQA datasets is no easy. Thus, learning-based algorithms utilizing only regression loss are further developed by researchers. Learning to rank affords us with approaches taking better use of labeled IQA data. These approaches judge the rank of predicted results with corresponding ground-truths in pair-wise comparisons, without hardly forcing them to be the same\cite{41, 42}. This technique is applied widely in recommendation and searching algorithms for a few years, but the information of rankings is first introduced by RankIQA\cite{23} to the NR-IQA tasks. The authors also point out that pair-wise rank loss can be more efficient when using batched training strategy by sequentially calculating the loss between each image pair in the batches. RankIQA has achieved good performance in publicly released datasets at that time. Later works about learning to rank makes great effort in designing losses\cite{43, 44}, which also promote the study of ranking-based IQA algorithms. 

\subsection{IQA based on vision transformers}

The transformer takes the form of encoder-decoder architecture and employs multi-head attention to gather information from different representation subspaces and positions\cite{27}. It has already gained huge success in the field of natural language processing(NLP)\cite{45,46,47}, which also inspires the application of transformer architecture in vision tasks. Recent works in this area \cite{26,27,28, 48, 49}demonstrate its excellent performance in solving computer vision problem. Among these, One of the most typical models is the Vision Transformer\cite{27, 48}. ViT takes cropped image patches as sequential inputs, and designed a position embedding to model the relationship between different patches. Another significant improvement is Swin Transformer\cite{31}. it implements the hierarchy architecture with shifted windows to improve both accuracy and computation efficiency. Previous works of transformer-based IQA algorithms are usually modified in the scheme of ViT\cite{29,30,50,51,52, 54}. TRIQ\cite{29} takes CNN features as the input of transformer encoder and utilizes a position embedding with fixed length to gather the information across different positions. MUSIQ\cite{30} presents a multi-scale IQA transformer to capture perceptual quality at different granularities. TranSLA\cite{52} incorporates an auxiliary network to predict saliency maps, which forces the main network to pay more attention to salient areas. These works have achieved even superior performances than the CNN-based ones. In the following part of this paper, we will also compare the results of these methods on publicly available datasets with our proposed method.

\section{Proposed method}
\begin{figure*}[h]
    \centering
    \includegraphics[scale = 0.36]{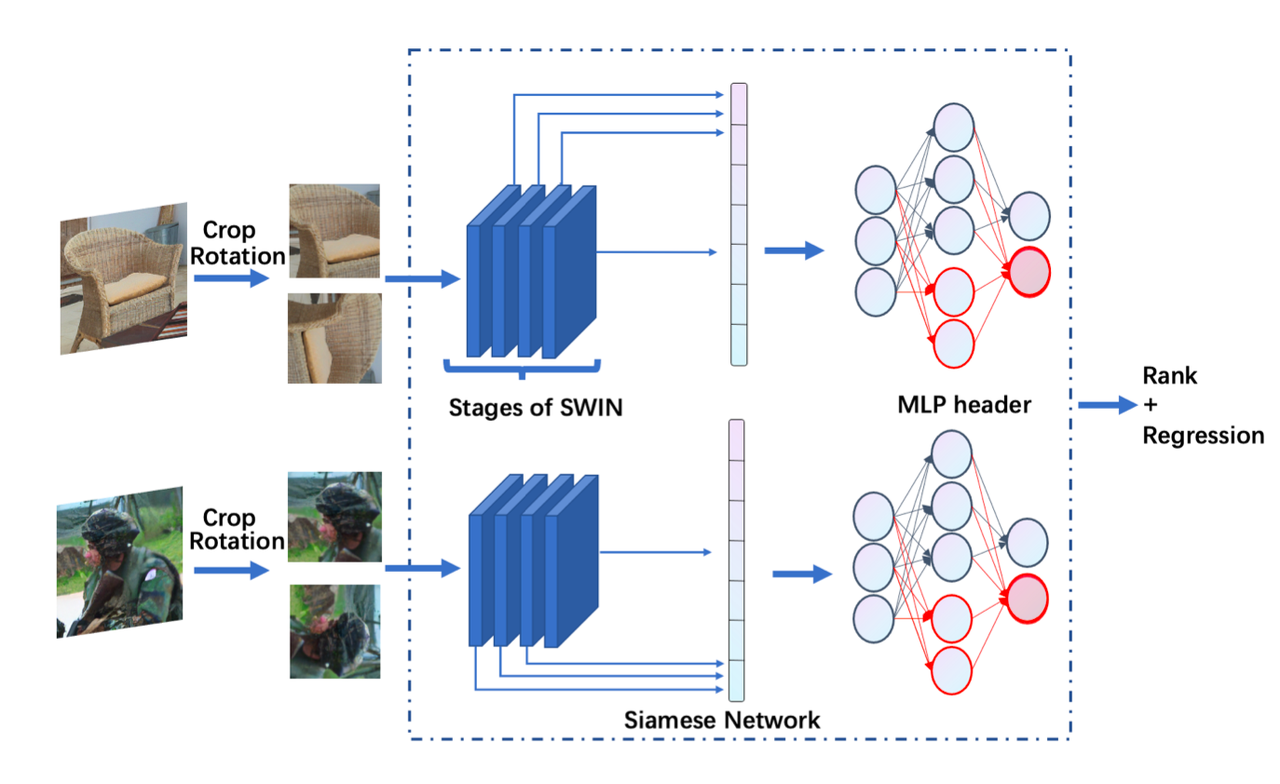}
    \caption{Multi-Stage fusing Swin transformer based on Siamese Network}
    \label{fig:swin}
\end{figure*}

In this paper, we have proposed an no-reference IQA algorithm based on Swin Transformer with fused features from multiple stages. Fig.\ref{fig:swin}  illustrates the overall architecture of our proposed model. It is composed by a fusing stage Swin Transformer backbone to extract quality features and a multi-layer fully connected header to predict the score. Cropped patches from resized images serve as the inputs of the network. Then we concatenate features from different stages together before feeding them into the prediction head. Rank and regression loss are combined together to perform model optimization. In the meantime, Siamese networks with shared parameters are employed to accommodate the pair-wise training strategies. The tested images are cropped after augmentations such as random rotation and color space changing.  These tricks also bring significant improvements to our model performance.
\subsection{Multi-Stage fusing Swin Transformer based Model}
To extract quality features of different scales, we adopt the Swin Transformer as our backbone and modify it with feature fusion from multiple stages. The Swin Transformer integrates the advantages of CNNs with that of vision transformers, which helps us achieve a better trade-off between accuracy and efficiency in IQA tasks. The model framework is represented in Fig.\ref{fig:swin}.
The distorted images are  rotated and resized to a fixed resolution. Then, the resized RGB images will be split into non-overlapping patches by a path splitting module\cite{31}. These patches are treated as "token". The token features are defined as raw RGB values. In our case, we use 4x4 as the patch size, hence the feature dimension of each patch is 4×4×3=48. Swin Transformer blocks\cite{31} will process these patches as tokens with a total number of $\frac{H}{4}*\frac{W}{4}$ afterwards, where H and W represent the height and width of input image. We fuse four stage outputs of Swin Transformer and concatenate them together to aggregate hierarchical information of the input image. The output size of each Swin Transformer stage  is $\frac{H}{4}*\frac{W}{4}*C,\frac{H}{8}*\frac{W}{8}*2C,\frac{H}{16}*\frac{W}{16}*4C,\frac{H}{32}*\frac{W}{32}*8C
$ where $C$ denotes arbitrary dimension processed by linear embedding. After concatenating and flattening the outputs of four stages, a fused feature map with size  $\frac{H}{4}*W*C$ is obtained.A MLP header is used to map obtained feature map to image quality score. Two fully connected(FC) layers are used in the MLP module. The first FC layer is followed by GELU activation, and the last layer will generate final score of given distorted image.

Our model adopt the Siamese networks which trains two images simultaneously to learn from image rankings.\cite{23} The two networks are sharing weights during the overall training process. Pairs of images and MOS labels are sent into the model. The outputs of the Siamese networks are sent to the loss module. As mentioned before, the loss term used here combines rank loss with regression loss, where the rank loss guarantees the relative relationship of given image pairs. 
\subsection{Data augmentation}
\begin{figure}[h]
    \centering
    \includegraphics[scale = 0.3]{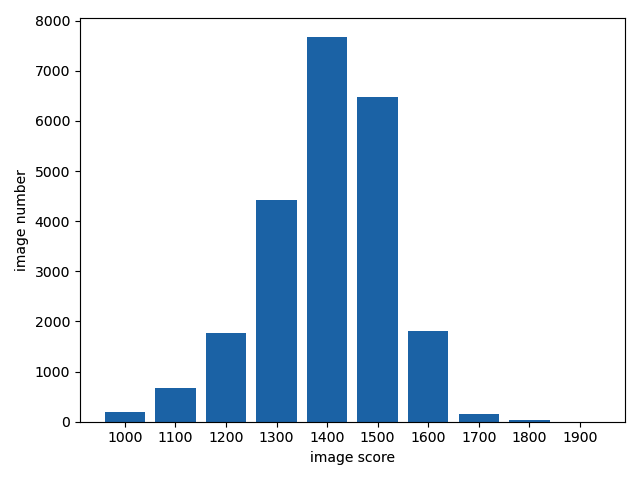}
    \caption{Label distribution of distorted images in PIPAL dataset}
    \label{fig:dist}
\end{figure}

Several data augmentation techniques are performed to increase the number of images in the training dataset and enhance robustness of our model. Firstly, we  resize images to a certain resolution and randomly crop patches of images for training.The label of each patch is simply set to be the same as the MOS of original image. Secondly, random rotations and colorspace changes are employed to the cropped patches. In the training time, the patches will be changed to one of these four color spaces at random: RGB, HSV, LAB and grayscale. While in the test phase, these patches will be kept in original RGB. Besides, for PIPAL dataset suffering from extremely unbalanced data, we use a weighted sampling strategy to get a more balanced distribution. The histogram in Fig.\ref{fig:dist} shows the number of image distributions with respect to their MOSs. It can be seen that the number of images with low scores and high scores are taking relatively small portions of the total images, hence images with these scores will be up-sampled in each batch by the weighted sampling strategy .
The test time augmentations(TTAs) are also performed on the test dataset. For TID2013 and KonIQ-10k datasets, five-crop was used on the resized input images and harmonic mean is calculated on the patch scores. The formula of harmonic mean is presented as below, where $n$ represents total number of patches from a given image and $x_i$ denotes the i-th image patch score predicted by our model.
\begin{equation}
harmonic\_mean = \frac{n}{\Sigma\frac{1}{x_i}}
\end{equation}
For PIPAL dataset, we use N random crops from every test image and a harmonic mean of patch scores is calculated as the final prediction. Setting N to a larger number brings better performance on test data of PIPAL, but the improvements are very small while N is too large.

%-------------------------------------------------------------------------
\subsection{Pairwise training }
With ideas borrowed from learning to rank\cite{41, 42, 60} and ranking-based IQA algorithms\cite{23, 60}, we implement the ranking constraint together with the regression loss during the training process. A loss function with two terms, including both ranking and regression, is utilized to jointly optimize the model:
\begin{equation}
loss_{total} = loss_{reg} + loss_{rank}
\end{equation}
For the regression task, we use Euclidean loss, which is given by:         
\begin{equation}               
loss_{reg} = \cfrac{1}{2N}
\sum_{
\begin{subarray}{l}
i= 1:N
\end{subarray}}
 {\vert{\hat{y_i} - y_i}\vert}^2
\end{equation}
where$y_i$is the ground-truth score for image-i, and $\hat{y_i}$ is the estimated score by our model. The regression loss is calculated between the prediction and the ground truth of each image to evaluate the absolute distance between them. 
For the ranking task, we adopt a pair-wise ranking loss\cite{60} to explicitly exploit relative rankings of image pairs in the PIPAL dataset. The ranking loss we employ here is in an exponential form, addressing the different importance of pairs with small and large ranking errors. The ranking loss is given by: 
\begin{equation}                      
loss_{rank} = \cfrac{2}{N}\sum_{
\begin{subarray}
{l}
i=0:2:N   
\end{subarray}}
\begin{cases}
e^{\hat{y}^i-\hat{y}^{i+1}},   &\text{if } {y^i<y^{i+1}}\\
0, &\text{others } 
\end{cases}
\end{equation}
The simultaneous applications of these two loss terms provide knowledge of both absolute ground truths and relative rankings for model training. The use of either alone can not reach the performance of the combined loss, which will be proved by ablation studies in the next section.
\section{Experiments}
\subsection{ Datasets and evaluation metrics }

The performance of our proprosed method are evaluated on three publicly available NR-IQA datasets. The first one is a synthetically distorted dataset, which is TID2013 \cite{56} with 25 reference images and 3000 distorted images. The distortions of TID2013 images are mostly traditional ones, such as quantization noise, gaussian blur, compression,  artifacts caused by image denoising and so on, while the MOSs are obtained in laboratory environments. The second is an authentically distorted dataset, KonIQ-10k \cite{55}. It has 10073 images with MOS. Most images in KonIQ-10k are generated in th wild.  Detailed comparisons are made  on these two datasets between the proposed method and existing NR-IQA algorithms, with results demonstrated in Section 4.3. Another dataset is PIPAL\cite{57}, which is released by NTIRE 2022 Perceptual Image Quality Assessment Challenge. Distoritions of PIPAL images are quite different from the aforementioned ones, which can be induced by traditional and GAN-based image enhancement algorithms, or enven a mixture of both\cite{53}. The ground truth labels of validation and test data in PIPAL dataset are not provided, so we only compare the results announced by the challenge organizers in the final testing phase.
Here we employ two commonly used evaluation metrics in performance comparison: Spearman’s Rank-order Correlation Coefficient (SRCC) and Pearson’s Linear Correlation Coefficient (PLCC). SRCC measures the monotonic relationship between the predicted results and the ground truths, which is computed as:
\begin{equation}
SRCC = 1- \frac{6\Sigma_{i=1}^{N}d_{i}^{2}}{N(N^{2}-1)}
\end{equation}
where $d_i$is the distance between rank orders in predictions and in MOS labels of the same image, $N
$ is number of images. Slightly different from SRCC, PLCC measures the linear correlation of predictions and ground truths:
\begin{equation}
PLCC = \frac{\Sigma_{i=1}^{N}(s_{i}-\bar{s})(p_{i} - \bar{p})}{\sqrt{\Sigma_{i=1}^{N}(s_{i}-\bar{s})^{2}\Sigma_{i=1}^{N}(p_{i} - \bar{p})^{2}}}
\end{equation}
where $p_i$ and $s_i$are the predicted score and MOS of each image respectively

\subsection{Implementation details}

The proposed model was implemented by PyTorch framework with model training conducted on 4 TESLA V100 GPUs. Instead of training from scratch, we employed pretrained model on ImageNet and finetuned it on IQA datasets. It takes about 1 hour for each round of training. For TID2013 and KonIQ-10k, we randomly sampled 80\% images for training and 20\% for testing like previous works do\cite{23, 25, 29, 52}. For PIPAL, we simply used the provided training data and results were tested on the online server.In the training phase, the input image will be resized to a certain resolution and a patch of fixed size will be cropped from the resized image. For experiments on TID2013 and Koniq10k, which have higher resolution images, the size of the patch fed into model is set as 384 × 384. When it comes to PIPAL with image resolution of 288 × 288, we set a smaller patch size of 224 × 224. The cropped patches will then be augmented and batched by data pre-processing procedure during the process of training. In the testing phase, image patches are also acquired from the given image. We randomly extract N overlapping patches and the final quality score is calculated by the  harmonic mean of N individual patch scores.
The training is conducted using an ADAMW optimizer with paramers $\beta1 = 0.9, \beta2 = 0.999$. The batch size of every single GPU is set to 64. Parallel training is performed on multiple GPUs to accelerate learning with large batch size. The basic learning rate is initialized as $2e^{-5} $ and a cosine learning strategy is used with a few warm-up epochs at the beginning. 

\subsection{Experimental results}

\begin{figure*}[h]
    \centering
    \includegraphics[scale = 0.3]{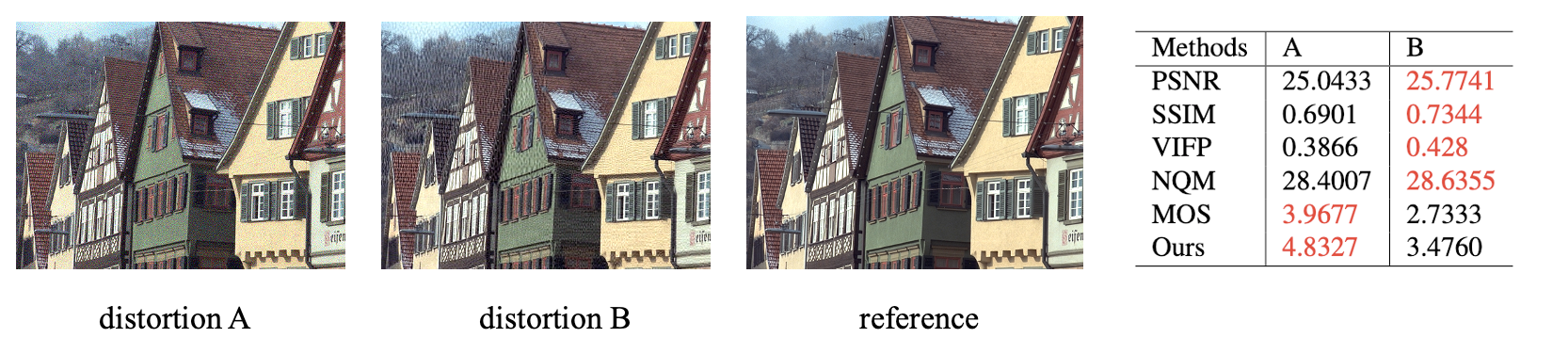}
    \caption{Typical comparison case (choose from TID2013\cite{56}) of image pairs with different quality scores from different algorithms.}
    \label{fig:vision}
\end{figure*}

\begin{figure}[h]
    \centering
    \includegraphics[scale = 0.15]{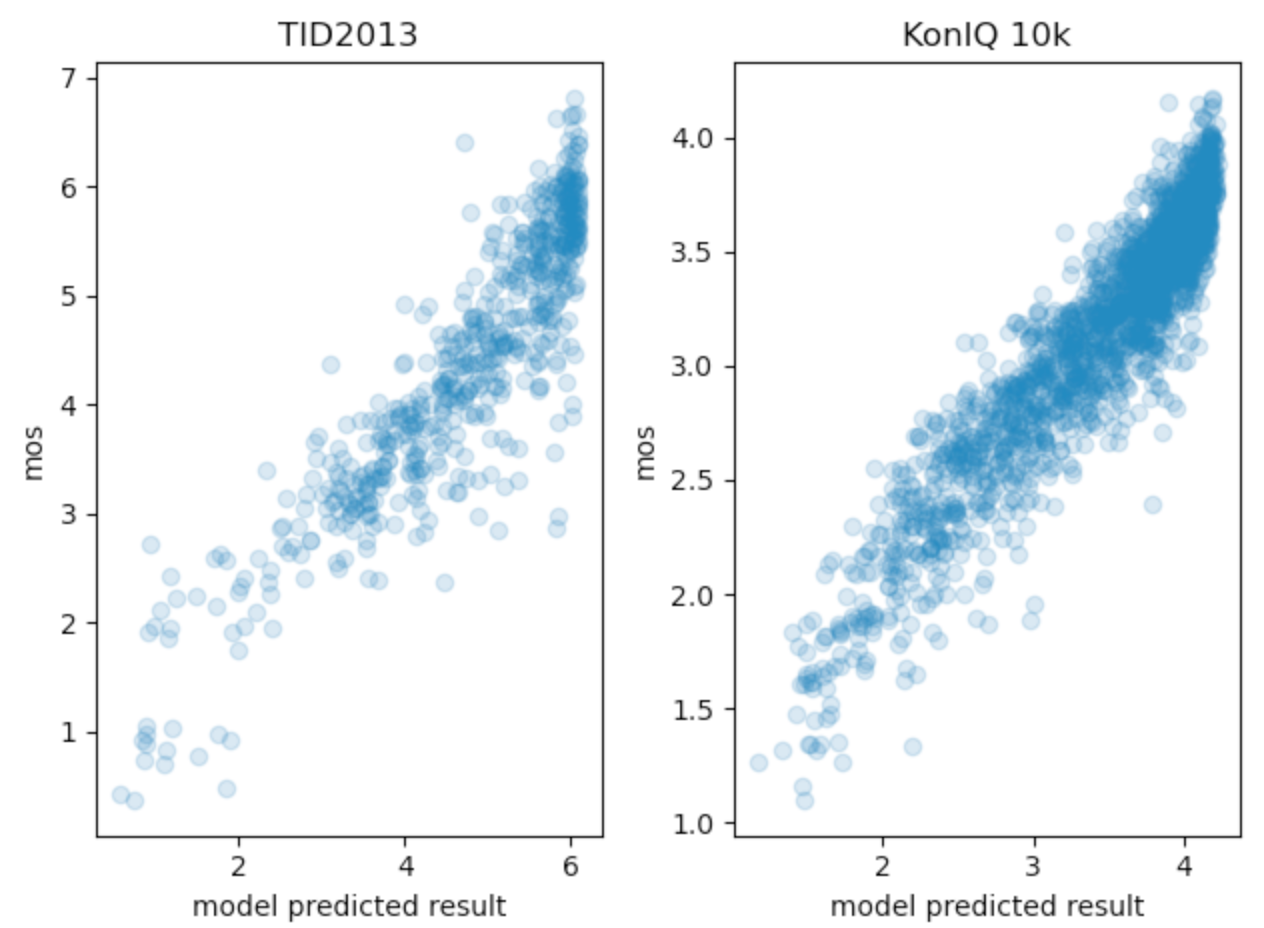}
    \caption{Scalar plots of ground-truth scores against the predicted scores of proposed IQA model finetuned on TID2013 and KonIQ 10k.}
    \label{fig:plot}
\end{figure}

The benchmark results of the existing IQA methods on TID2013  and  KonIQ-10k  are shown in Table \ref{tab:datasets_compare}, with results given by the proposed algorithm listed at the bottom. These models are trained and tested on each dataset respectively. Some results are borrowed from the original papers\cite{16,25,54,58,59}. This table demonstrates that our method outperforms recent developed algorithms with competitive performance,  in terms of both PLCC and SRCC metrics. It proves to be a promising approach in the field of quality assessment on different datasets with both syntheticand authentic distortions. The scalar plots of ground truth MOSs and model predicted results of different datasets are shown in Fig.\ref{fig:plot}. It can be seen that the predicted scores of our proposed method show good consistency with the human-rated MOSs, which means it tends to give a higher score when the perceptual image quality is higher.
Figure \ref{fig:vision} shows a typical case that the image pairs have obtained different quality scores from different algorithms. The characters in red indicate the image of higher quality. It appears that the predicted scores of our proposed method show better consistency with the MOSs.

\begin{table}[h]
\begin{tabular}{l|ll|ll}
 \centering
Datasets          & \multicolumn{2}{l|}{KonIQ-10k}  & \multicolumn{2}{l}{TID2013}       \\ \hline
Methods           & PLCC           & SRCC           & PLCC            & SRCC            \\ \hline
BRISQUE {[}16{]}  & 0.681          & 0.665          & {0.610}  & {0.544}  \\
HyperIQA {[}25{]} & 0.917          & 0.906          & 0.858           & 0.840           \\
MetaIQA{[}58{]}   & 0.887          & 0.850          & 0.868           & 0.856           \\
FPR{[}59{]}       & 0.901          & 0.899          & 0.887           & 0.872           \\
TReS{[}54{]}      & 0.928          & 0.915          & 0.883           & 0.863           \\
\textbf{Ours}     & \textbf{0.954} & \textbf{0.946} & \textbf{0.895} & \textbf{0.882} \\ \hline
\end{tabular}
  \caption{Overall performance evaluation on two image databases.}
  \label{tab:datasets_compare}
\end{table}

Results produced by different  algorithms on PIPAL dataset are listed in table \ref{tab:pipal_compare} . Besides the leading results on the other two standard datasets, the performance of proposed method on this dataset again surpasses those of the available state-of-the-art algorithms, which shows a strong generalizability of the algorithm for tackling similar no-reference IQA problems.

\begin{table}[]
\centering
\begin{tabular}{l|lll}
\hline
IQA Name            & MainScore                  & SRCC                       & PLCC  \\ \hline
PSNR \cite{61}        & \multicolumn{1}{l|}{0.572} & \multicolumn{1}{l|}{0.269} & 0.303 \\
SSIM \cite{62}   & \multicolumn{1}{l|}{0.785} & \multicolumn{1}{l|}{0.377} & 0.407 \\
LPIPS-Alex \cite{63} & \multicolumn{1}{l|}{1.176} & \multicolumn{1}{l|}{0.584} & 0.592 \\
FSIM \cite{64}       & \multicolumn{1}{l|}{1.138} & \multicolumn{1}{l|}{0.528} & 0.610  \\
NIQE \cite{17}      & \multicolumn{1}{l|}{0.142} & \multicolumn{1}{l|}{0.030} & 0.112 \\
MA \cite{65}       & \multicolumn{1}{l|}{0.398} & \multicolumn{1}{l|}{0.174} & 0.224 \\
PI \cite{66}         & \multicolumn{1}{l|}{0.276} & \multicolumn{1}{l|}{0.123} & 0.153 \\
Brisque \cite{16}    & \multicolumn{1}{l|}{0.184} & \multicolumn{1}{l|}{0.087} & 0.097 \\
\textbf{Ours}       & \multicolumn{1}{l|}{\textbf{1.437}} & \multicolumn{1}{l|}{\textbf{0.737}} & \textbf{0.700} \\ \hline
\end{tabular}
  \caption{Benchmark results on PIPAL dataset\cite{67,68}.}
  \label{tab:pipal_compare}
\end{table}

\subsection{Ablation Studies}
Ablation studies are carried out to analyze the effectiveness of the proposed algorithm. 
We have performed experiments on the network architecture. Table \ref{tab:fusion_compare} shows the results of models with and without the feature fusion module. It indicates that models with fused features have better performance. This is in accordance to our previous analysis as these models obtained more information from different scales.
\begin{table}[]
\centering
\begin{tabular}{l|ll}
\hline
Datasets            & \multicolumn{2}{l}{KonIQ-10k}        \\ \hline
Methods             & \multicolumn{1}{l|}{PLCC}   & SRCC   \\ \hline
Ours without feature fusion & \multicolumn{1}{l|}{0.953}  & 0.942  \\
\textbf{Ours with feature fusion}    & \multicolumn{1}{l|}{\textbf{0.954}} & \textbf{0.946} \\ \hline
\end{tabular}
  \caption{Model with and without feature fusion module test on KonIQ-10k dataset.}
  \label{tab:fusion_compare}
\end{table}
The effects of the rank and regression loss terms are also analyzed. Table \ref{tab:loss_compare} shows the results of models trained with different loss terms: regression loss, rank loss and a combination of both. The sole utilization of regression and rank loss has both achieved poorer performance compared with the joint training strategy. It illustrates the necessity of simultaneously optimizing the model with both information.
\begin{table}[]
\centering
\begin{tabular}{l|l|l}
\hline
Methods           & PLCC            & SRCC            \\ \hline
mse               & 0.951         & 0.940          \\
rank              & 0.941         & 0.945          \\
\textbf{mse+rank} & \textbf{0.954} & \textbf{0.946} \\ \hline
\end{tabular}
  \caption{Overall performance evaluation with different loss functions on KonIQ-10k.}
  \label{tab:loss_compare}
\end{table}

\subsection{NTIRE 2022 Perceptual IQA Challenge Track 2 No-reference}
For NTIRE 2022 Perceptual Image Quality Assessment Challenge, we employed essentially the same model as the aforementioned one. In addition, an ensembling strategy was performed to improve the robustness of model,  and prevent the occurrence of both over-fitting and under-fitting. The ensembling is done by averaging the results of models with different sizes of resized input images, different types of backbones and also different combinations of augmentations. In detail, the final result is averaged from four models, which are: Base-224, Large-256 and two Large-288. The ’Base’ and ’Large’ indicate the type of backbones, and the numbers after them indicate the size of resized images. Patch size is kept 224 × 224. These four models are ensembled together to further improve the prediction.
The results of the challenge are shown in Table \ref{tab:ntire_compare}. We got second place in the testing phase and the MainScore is very close to the top method. 
\begin{table}[]
\centering
\begin{tabular}{l|l|ll}
\hline
Teams & MainScore & PLCC                       & SRCC  \\ \hline
1st    & 1.444     & \multicolumn{1}{l|}{0.740} & 0.704 \\
DTIQA  & 1.437     & \multicolumn{1}{l|}{0.737} & 0.700 \\
3rd    & 1.422     & \multicolumn{1}{l|}{0.725} & 0.697 \\
4th    & 1.407     & \multicolumn{1}{l|}{0.726} & 0.681 \\
5th    & 1.390     & \multicolumn{1}{l|}{0.720} & 0.671 \\ \hline
\end{tabular}
  \caption{Result of NTIRE 2022 No-Reference IQA Challenge in testing phase.}
  \label{tab:ntire_compare}
\end{table}
%-------------------------------------------------------------------------
\section{Conclusion}
In this paper, we proposed a method to assess perceptual image quality based on the architecture of Swin Transformer. The method was first developed for the NTIRE 2022 Perceptual IQA Challenge, but also proved effective when tested on other IQA datasets with diverse types of distortions. A unified loss function was used to train the model using prior knowledge from both absolute ground truths and relative rankings. Data augmentations and training tricks were employed to improve performance. The model takes randomly cropped image patches as inputs, and predicts a score for every patch. A harmonic mean of output patch scores was calculated to represent the predicted quality of original image. The experiment results show the superior performance of the algorithm, thus it can be a promising method in solving NR-IQA problems.
%-------------------------------------------------------------------------

%%%%%%%%% REFERENCES


\begin{thebibliography}{1}\itemsep=-1pt

\bibitem{Gu_2021_CVPR}
Jinjin Gu, Haoming Cai, Chao Dong, Jimmy~S. Ren, Yu Qiao, Shuhang Gu, and Radu
  Timofte.
\newblock Ntire 2021 challenge on perceptual image quality assessment.
\newblock In {\em Proceedings of the IEEE/CVF Conference on Computer Vision and
  Pattern Recognition (CVPR) Workshops}, pages 677--690, June 2021.

\bibitem{gu2022ntire}
Jinjin Gu, Haoming Cai, Chao Dong, Jimmy~S. Ren, Radu Timofte, et~al.
\newblock {NTIRE} 2022 challenge on perceptual image quality assessment.
\newblock In {\em Proceedings of the IEEE/CVF Conference on Computer Vision and
  Pattern Recognition (CVPR) Workshops}, 2022.

\end{thebibliography}


\begin{thebibliography}{9}
\bibitem{1}
 H. Bovik. Image information and visual quality. IEEE Trans. Image Process., 5(2):430, 2006.
 
 \bibitem{2}
W. Hou, X. Gao, D. Tao, et al. Blind Image Quality Assessment via Deep Learning. IEEE Transactions on Neural Networks and Learning Systems, 26(6):1275-1286, 2017.

 \bibitem{3} 
 W. Zhou, A. C. Bovik and L. Lu. Why is image quality assessment so difficult? IEEE ICASSP, 3313-3316, 2002
  
  \bibitem{4} 
D. Chen, Y. Wang, W. Gao, et al. No-Reference Image Quality Assessment: An Attention Driven Approach. IEEE Trans. Image Process., 29(99):6496-6506, 2020.
 
  \bibitem{5} 
W. Zhou, Z. Wang and Z. Chen. Image Super-Resolution Quality Assessment: Structural Fidelity Versus Statistical Naturalness. arXiv preprint arXiv:2105.07139, 2021.
 
  \bibitem{6} 
J. C. Mier, E. Huang and H. Talebi, et al. Deep Perceptual Image Quality Assessment for Compression. arXiv preprint arXiv:2103.01114, 2021.
 
   \bibitem{7} 
S. Athar, Z. Wang, et al. Deep Neural Networks for Blind Image Quality Assessment: Addressing the Data Challenge. arXiv preprint arXiv:2109.12161, 2021.
 
   \bibitem{8} 
P. Ye. No-Reference Image Quality Assessment Using Visual Codebooks. IEEE Trans. Image Process., 21(7):3129-3138, 2012.
 
   \bibitem{9} 
H. Liu and I. Heynderickx. Visual Attention in Objective Image Quality Assessment: Based on Eye-Tracking Data. IEEE Trans. Circuit Syst. for Video Technol, 21(7):971-982, 2011.
 
   \bibitem{10} 
W. Zhang, K. Ma, G. Zhai, Yang X. Learning to blindly assess image quality in the laboratory and wild. In IEEE Int. Conf. Image Process., 111-115, 2020.
 
  \bibitem{11} 
Z. Ying, M. Mandal, D. Ghadiyaram, et al. Patch-VQ: 'Patching Up' the Video Quality Problem. arXiv preprint arXiv:2011.13544, 2020.
 
   \bibitem{12} 
W. Sun, Q. Liao, J. H. Xue, et al. SPSIM: A superpixel-based similarity index for full-reference image quality assessment. IEEE Trans. Image Process., 27(9): 4232-4244, 2018.
 
   \bibitem{13} 
B. Chen, L. Zhu, C. Kong, et al. No-Reference Image Quality Assessment by Hallucinating Pristine Features. arXiv preprint arXiv: 2108.04165, 2021.
 
   \bibitem{14} 
S. Ahn, Y. Choi, K. Yoon. Deep learning-based distortion sensitivity prediction for full-reference image quality assessment. Proceedings of the IEEE Conf. Comput. Vis. Pattern Recog., 344-353, 2021.
 
    \bibitem{15} 
K. Rahul, A. K. Tiwari. FQI: feature-based reduced-reference image quality assessment method for screen content images. IET Image Processing, 13(7): 1170-1180, 2019.
 
    \bibitem{16} 
A. Mittal, A. K. Moorthy, A. C. Bovik. No-Reference Image Quality Assessment in the Spatial Domain. IEEE Trans. Image Process., 21(12):4695, 2012.
 
    \bibitem{17} 
A. Mittal. Making a 'Completely Blind' Image Quality Analyzer. IEEE Signal Processing Letters, 20(3):209-212, 2013.
 
    \bibitem{18} 
O. I. Abiodun, A. Jantan, A. E. Omolara, et al. State-of-the-art in artificial neural network applications: A survey. Heliyon, 4(11), 2018.
 
    \bibitem{19} 
A. Krizhevsky, I. Sutskever, G. E. Hinton. ImageNet classification with deep convolutional neural networks. Advances in neural information processing systems, 25, 2012.
 
    \bibitem{20} 
K. He, X. Zhang, S. Ren, et al. Deep residual learning for image recognition. Proceedings of the IEEE Conf. Comput. Vis. Pattern Recog., 770-778, 2016.
 
    \bibitem{21} 
S. Minaee, N. Kalchbrenner, E. Cambria, et al. Deep learning-based text classification: a comprehensive review. ACM Computing Surveys (CSUR), 54(3): 1-40, 2021.
 
    \bibitem{22} 
K. Le, Y. Peng, L. Yi , et al. Convolutional Neural Networks for No-Reference Image Quality Assessment. IEEE Conf. Comput. Vis. and Pattern Recog., IEEE, 2014.
 
    \bibitem{23} 
X. Liu, J. Weijer, A. D. Bagdanov. RankIQA: Learning from Rankings for No-Reference Image Quality Assessment. IEEE Int. Conf. Comput. Vis., 1040-1049, 2017.
 
    \bibitem{24} 
W. Zhang, K. Ma and J. Yan, et al. Blind image quality assessment using a deep bilinear convolutional neural network. IEEE Trans. Circuits Syst. Video Technol, 30(1):36–47, 2018.
 
    \bibitem{25} 
S Su, Q. Yan, Y Zhu, et al. Blindly assess image quality in the wild guided by a self-adaptive hyper network. In Proceedings of the IEEE Conf. Comput. Vis. Pattern, 3667–3676, 2020.
 
    \bibitem{26} 
X. Zhu, W. Su, et al. Deformable DETR: Deformable transformers for end-to-end object detection. arXiv preprint arXiv:2010.04159, 2020.
 
    \bibitem{27} 
A. Dosovitskiy, L. Beyer, A. Kolesnikov A, et al. An image is worth 16x16 words: Transformers for image recognition at scale. arXiv preprint arXiv:2010.11929, 2020.
 
    \bibitem{28} 
H. Chen, Y. Wang, et al. Pre-trained image processing transformer. InProceedings of the IEEE Conf. Comput. Vis. Pattern, 12299-12310, 2021.
 
    \bibitem{29} 
J. You, J. Korhonen. Transformer for image quality assessment. In2021 IEEE Int. Conf Image Process., 1389-1393, 2021.
 
    \bibitem{30} 
J. Ke, Q. Wang, Y. Wang, et al. Musiq: Multi-scale image quality transformer. In Proceedings of the IEEE Int. Conf. Comput. Vis., 5148-5157, 2021.
 
    \bibitem{31} 
Z. Liu, Y. Lin, Y. Cao, H. Hu, et al. Swin transformer: Hierarchical vision transformer using shifted windows. In Proceedings of the IEEE Int. Conf. Comput. Vis., 10012-10022, 2021.
 
    \bibitem{32} 
J. Liang, J. Cao, G. Sun, K. Zhang, et al. SwinIR: Image restoration using swin transformer. In Proceedings of the IEEE Int. Conf. Comput. Vis., 1833-1844, 2021.
 
    \bibitem{33} 
N. Ponomarenko, O. Ieremeiev, V. Lukin, et al. Color image database TID2013: Peculiarities and preliminary results. In European workshop on visual information processing, 106-111, 2013.
 
    \bibitem{34} 
R. A. Manap, L. Shao. Non-distortion-specific no-reference image quality assessment: A survey. Information Sciences, 301: 141-160, 2015.
 
    \bibitem{35} 
A. K. Moorthy and A. C. Bovik. A two-step framework for constructing blind image quality indices. IEEE Signal processing letters, 17(5): 513-516, 2010.
 
    \bibitem{36} 
N. Venkatanath, D. Praneeth, et al. Blind image quality evaluation using perception based features. IEEE National Conference on Communications, 1-6, 2015.
 
    \bibitem{37} 
J. Xu, P. Ye and Q. Li, et al. Blind Image Quality Assessment Based on High Order Statistics Aggregation. IEEE Trans. Image Process., 25(9):4444-4457, 2016.
 
     \bibitem{38} 
S. Bosse, D. Maniry, et al. Deep neural networks for no-reference and full-reference image quality assessment. IEEE Trans. Image Process., 27(1): 206–219, 2017.
 
     \bibitem{39} 
W. Zhang, K. Ma and G. Zhai, et al. Uncertainty-aware blind image quality assessment in the laboratory and wild. IEEE Trans. Image Process., 30: 3474-3486, 2021.
 
     \bibitem{40} 
C. Li, A. Bovik, and X. Wu. Blind image quality assessment using a general regression neural network. IEEE Transactions on Neural Networks, 22(5):793–799, 2011.
 
     \bibitem{41} 
W. Chen, T.-Y. Liu and Y. Lan, et al Ranking measures and loss functions in learning to rank. In Advances in Neural Information Processing Systems, 315–323, 2009.
 
      \bibitem{42} 
M. Li, X. Liu, et al. Learning to rank for active learning: A listwise approach. Int. Conf. Pattern Recog, 5587-5594, 2021.
 
      \bibitem{43} 
S. Bruch. An alternative cross entropy loss for learning-to-rank. Proceedings of the Web Conference, 118-126, 2021.
 
      \bibitem{44} 
Minghan Li, Xialei Liu, Joost van de Weijer, and Bogdan Raducanu. Learning to rank for active learning: A listwise approach. In 25th International Conference on Pattern Recognition, ICPR 2020, Virtual Event / Milan, Italy, January 10-15, 2021, pp. 5587–5594. IEEE, 2020.
 
      \bibitem{45} 
J. Devlin, M.-W. Chang, and K. Lee, et al. Bert: Pre-training of deep bidirectional transformers for language understanding. arXiv preprint arXiv:1810.04805, 2018.
 
      \bibitem{46} 
Z. Yang, Z. Dai Z and Y. Yang, et al. XLNet: Generalized autoregressive pretraining for language understanding. Advances in neural information processing systems, 32, 2019.
 
      \bibitem{47} 
C. Raffel, N. Shazeer and A. Roberts, et al. Exploring the limits of transfer learning with a unified text-to-text transformer. arXiv preprint arXiv:1910.10683, 2019.
 
      \bibitem{48}
D. Zhou, B. Kang and X. Jin, et al. DeepVIT: Towards deeper vision transformer. arXiv preprint arXiv:2103.11886, 2021.
 
      \bibitem{49} 
K. Han, Y. Wang and H. Chen, et al. A survey on visual transformer. arXiv e-prints arXiv: 2012.12556, 2020.
 
      \bibitem{50} 
M. Cheon, S.-J. Yoon and B. Kang, et al. Perceptual image quality assessment with transformers. IEEE Conf. Comput. Vis. Pattern, 433–442, 2021.
 
      \bibitem{51} 
A. Chubarau, J.Clark. VTAMIQ: Transformers for Attention Modulated Image Quality Assessment. arXiv preprint arXiv:2110.01655, 2021.
 
      \bibitem{52} 
M. Zhu, G. Hou, and X. Chen,et al. Saliency-guided transformer network combined with local embedding for no-reference image quality assessment. IEEE Int. Conf. Comput. Vis., 1953–1962, 2021.
 
      \bibitem{53} 
J. Gu, et al. NTIRE 2021 challenge on perceptual image quality assessment. IEEE Conf. Comput. Vis. Pattern. 2021.
 
      \bibitem{54} 
S. Alireza Golestaneh, Saba Dadsetan, and Kris M. Kitani. No-reference image quality assessment via transformers, relative ranking, and selfconsistency. CoRR, 2108.06858, 2021.
 
      \bibitem{55} 
H. Lin, V. Hosu, and D. Saupe. Koniq-10k: Towards an ecologically valid and large-scale iqa database. CoRR, 2018. 
 
      \bibitem{56} 
N. Ponomarenko, L. Jin and O. Ieremeiev et al. Image database TID2013: Peculiarities, results and perspectives. Signal Processing: Image Communication, 2015.
 
      \bibitem{57} 
J. Gu, H. Cai and H. Chen, et al. PIPAL: A large-scale image quality assessment dataset for perceptual image restoration. In Eur. Conf. Comput. Vis., 633–651, 2020. 
 
      \bibitem{58} 
H. Zhu, L. Li and J. Wu, et al. MetaIQA: Deep metalearning for no-reference image quality assessment. IEEE Conf. Comput. Vis. Pattern, 14131–14140, 2020.
 
      \bibitem{59} 
B.-L. Chen, et al. No-Reference Image Quality Assessment by Hallucinating Pristine Features. arXiv preprint arXiv:2108.04165, 2021.
 
      \bibitem{60} 
L. Ma, L. Xu and Y. Zhang, et al. Ngan. No-reference retargeted image quality assessment based on pairwise rank learning. IEEE Trans. Multimedia, 18(11): 2228–2237, 2016.

\bibitem{61}
Group,V.Q.E.,etal.:Final report from the video quality experts group on the validation of objective models of video quality assessment. In: VQEG meeting, Ottawa, Canada, March, 2000 (2000).

\bibitem{62}
Wang,Z.,Bovik,A.C.,Sheikh,H.R.,Simoncelli,E.P.,etal.:Image quality assessment:from error visibility to structural similarity. IEEE transactions on image processing 13(4), 600–612 (2004).

\bibitem{63}
Zhang,R.,Isola,P.,Efros,A.A.,Shechtman,E.,Wang,O.:The unreasonable effectiveness of deep features as a perceptual metric. In: Proceedings of the IEEE Conference on Computer Vision and Pattern Recognition. pp. 586–595 (2018).

\bibitem{64}
Zhang,L.,Zhang,L.,Mou,X.,Zhang,D.:Fsim:A feature similarity index for image quality assessment. IEEE transactions on Image Processing 20(8), 2378–2386 (2011)	.

\bibitem{65}
Ma,C.,Yang,C.Y.,Yang,X.,Yang,M.H.:Learning a no-reference quality metric for single-image super-resolution. Computer Vision and Image Understanding 158, 1–16 (2017).

\bibitem{66}
Blau,Y.,Michaeli,T.:The perception-distortion trade off.In:ProceedingsoftheIEEEConference on Computer Vision and Pattern Recognition. pp. 6228–6237 (2018).

\bibitem{67}
Jinjin Gu, Haoming Cai, Chao Dong, Jimmy S. Ren, Yu Qiao, Shuhang Gu, and Radu Timofte. Ntire 2021 challenge on perceptual image quality assessment. In Proceedings of the IEEE/CVF Conference on Computer Vision and Pattern Recognition (CVPR) Workshops, pages 677–690, June 2021.

\bibitem{68}
Jinjin Gu, Haoming Cai, Chao Dong, Jimmy S. Ren, Radu Timofte, et al. NTIRE 2022 challenge on perceptual image quality assessment. In Proceedings of the IEEE/CVF Con- ference on Computer Vision and Pattern Recognition (CVPR) Workshops, 2022.

\end{thebibliography}
\end{document}